\documentclass[10pt,twocolumn,letterpaper]{article}
\usepackage{authblk}
\usepackage{wacv}              % To produce the CAMERA-READY version
\usepackage{graphicx}
\usepackage{amsmath}
\usepackage{amssymb}
\usepackage{booktabs}
\usepackage{subcaption}
\usepackage{multirow}
\usepackage{siunitx}
\usepackage{pifont}
\usepackage{soul} % Custom package
\usepackage[pagebackref,breaklinks,colorlinks]{hyperref}

\usepackage[capitalize]{cleveref}
\crefname{section}{Sec.}{Secs.}
\Crefname{section}{Section}{Sections}
\Crefname{table}{Table}{Tables}
\crefname{table}{Tab.}{Tabs.}

\def\wacvPaperID{2785} % *** Enter the WACV Paper ID here
\def\confName{WACV}
\def\confYear{2025}

\begin{document}

\title{A Pipeline and NIR-Enhanced Dataset for Parking Lot Segmentation}

% \author{
% Shirin Qiam \hspace{1.5cm} Saipraneeth Devunuri \hspace{1.5cm} Lewis J. Lehe\\
% University of Illinois at Urbana-Champaign\\
% Department of Civil and Environmental Engineering\\
% {\tt\small sqiam2@illinois.edu, sd37@illinois.edu, lehe@illinois.edu}
% }
\author{
    Shirin Qiam$^{1}$, Saipraneeth Devunuri$^{1}$, Lewis J. Lehe$^{1*}$\\
    $^1$Urban Traffic \& Economics Lab, University of Illinois at Urbana-Champaign\\
    $*$Corresponding Author\\
    {\tt\small \{sqiam2, sd37, lehe\}@illinois.edu}
}
% \author{
%     Shirin Qiam\thanks{Urban Traffic \& Economics Lab, University of Illinois at Urbana-Champaign.} \\
%     University of Illinois at Urbana-Champaign\\
%     {\tt\small sqiam2@illinois.edu}
%     \and 
%     Saipraneeth Devunuri\footnotemark[1] \\
%     {\tt\small sd37@illinois.edu}
%     \and 
%     Lewis J. Lehe\footnotemark[1] \\
%     {\tt\small lehe@illinois.edu} \\
% }

% \and
% Saipraneeth Devunuri\\
% University of Illinois at Urbana-Champaign\\
% Department of Civil and Environmental Engineering\\
% % First line of institution2 address\\
% {\tt\small sd37@illinois.edu}

% % First line of institution2 address\\
% {\tt\small lehe@illinois.edu}
% }
\maketitle

\begin{abstract}
    Discussions of minimum parking requirement policies often include maps of parking lots, which are time-consuming to construct manually.  Open-source datasets for such parking lots are scarce, particularly for US cities. This paper introduces the idea of using Near-Infrared (NIR) channels as input and several post-processing techniques to improve the prediction of off-street surface parking lots using satellite imagery. We constructed two datasets with 12,617 image-mask pairs each: one with 3-channel (RGB) and another with 4-channel (RGB + NIR). The datasets were used to train five deep learning models (OneFormer, Mask2Former, SegFormer, DeepLabV3, and FCN) for semantic segmentation, classifying images to differentiate between parking and non-parking pixels. Our results demonstrate that the NIR channel improved accuracy because parking lots are often surrounded by grass---even though the NIR channel needed to be upsampled from a lower resolution. Post-processing including eliminating erroneous ``holes,'' simplifying edges, and removing road and building footprints further improved the accuracy.  Best model, OneFormer trained on 4-channel input and paired with post-processing techniques achieves a mean Intersection over Union (mIoU) of 84.9\% and a pixel-wise accuracy of 96.3\%.
\end{abstract}

%%%%%%%%% BODY TEXT
\section{Introduction}
\label{sec:intro}

During the 20th century, nearly all US municipalities came to impose ``minimum parking requirements'' (MPRs) on new construction: mandates to provide parking in some proportion to the amount of floorspace or number of housing units proposed \cite{Ferguson2004}. Over the last twenty years, amid criticism that these requirements are harmful and based on faulty methodologies ~\cite{Shoup1999, Millard-Ball2015}, dozens of cities and several states have repealed or substantially liberalized their MPR's \cite{Lockhart2024}. Discussions of MPRs often involve estimates of how many parking spaces there are and how much land is devoted to parking \cite{Gabbe2021, Chester2015, Hoehne2019}. Nationally, it is estimated that between 0.64\% and 0.9\% of US land area (between 722 and 2010 million spaces) is parking \cite{Chester2011}.

\begin{figure}[ht]
    \centering
    \begin{subfigure}{0.22\textwidth}
        \centering
        \includegraphics[width=0.9\textwidth]{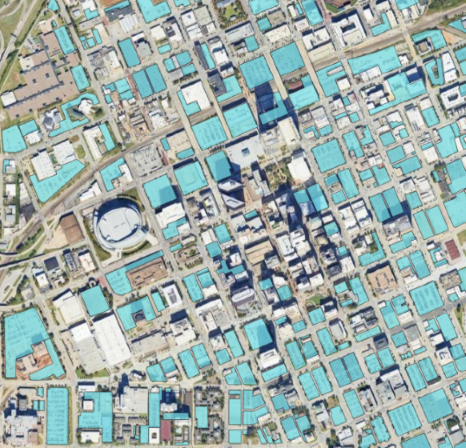}
        \caption{Tulsa, OK}
        \label{fig:tulsa}
    \end{subfigure}
    \begin{subfigure}{0.21\textwidth}
        \centering
        \includegraphics[width=0.9\textwidth]{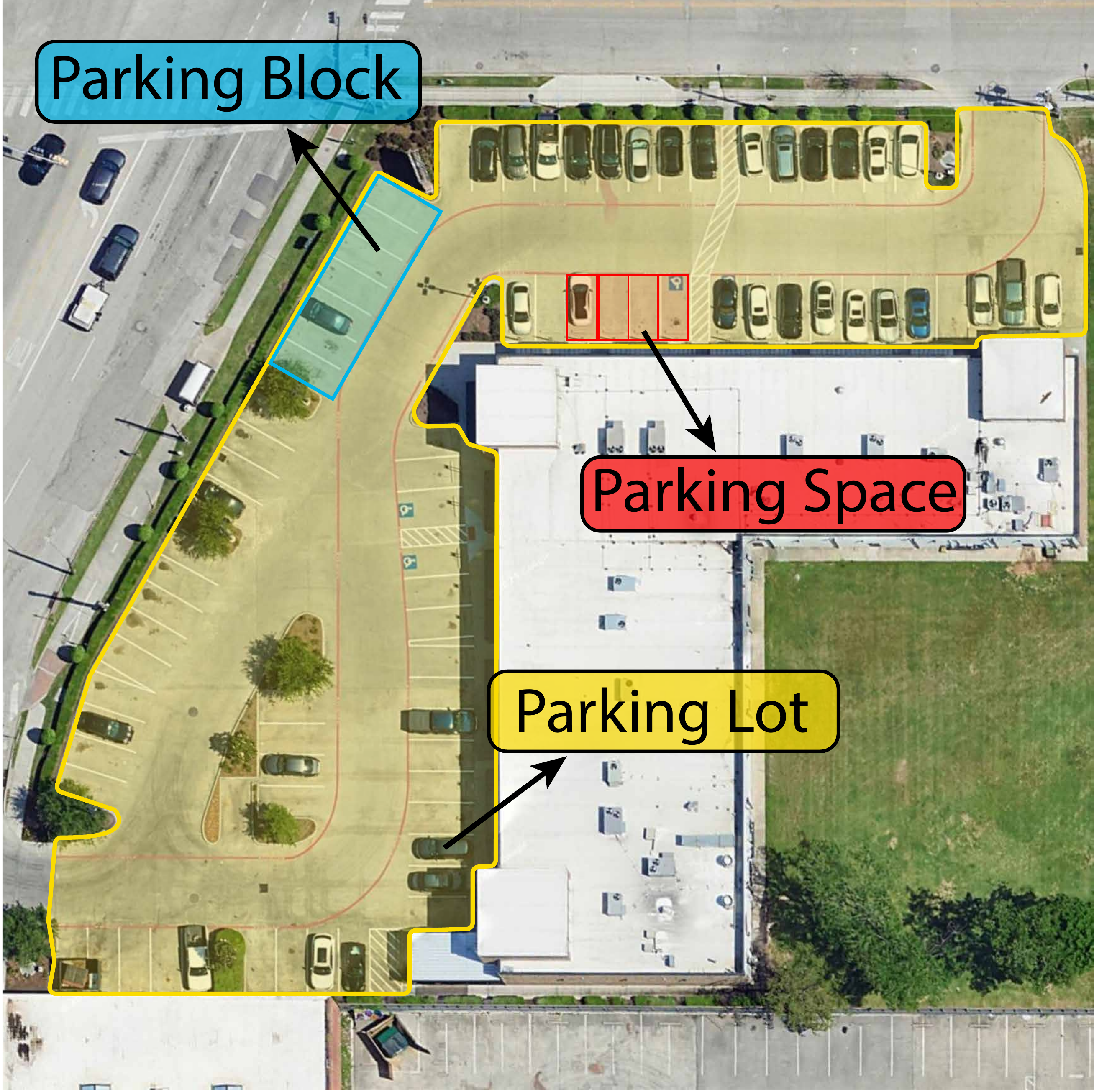}
        \caption{Annotation types}
        \label{fig:annotation_ex}
    \end{subfigure}
    \caption{Examples of parking lot annotations}
\end{figure}

In addition to these statistics, discussions of parking policy in media and legislatures have sometimes involved maps showing \emph{parking lots}---that is, the outlines of off-street parking lots (See Figure \ref{fig:tulsa}). Recently, a US-based group called the Parking Reform Network\footnote{The Parking Reform Network also assigned scores for cities to draw comparisons. The maps and scores for major cities can be viewed at \url{https://parkingreform.org/resources/parking-lot-map/}} has released interactive maps with parking lots drawn for the downtown areas of dozens of US cities. Such maps often draw significant attention in media wherever MPRs or other policies are being debated \cite{Fenton2023, Gann2023, Keatts2023, Oliver2023, Frolik2024}. However, the labor-intensive nature of manually creating these annotations has resulted in limited coverage, typically comprising only downtown areas of select cities. While the companies EarthDefine \cite{EarthDefine2021} and SafeGraph \cite{SafeGraph2024} sell datasets of parking lot annotations, there is a shortage of open-source datasets.

% \begin{figure}[ht]
%     \centering
%     \includegraphics[width=0.3\textwidth]{figures/annotation_ex.png}
%     \caption{Annotation example: parking lot boundaries}
%     \label{fig:annotation_ex}
% \end{figure}

This paper addresses the difficulty of constructing parking lot annotations for large areas. To this end, the following are the paper's main contributions:
\begin{itemize}
    \item An open-source dataset consisting of 12,617 satellite image/mask pairs of 512 x 512 dimensions. These masks outline $\sim$35,000 parking lots from 45 US cities\footnote{The dataset is available at \url{https://github.com/UTEL-UIUC/ParkSeg12k}.}.
    \item We employ five deep learning models (both CNN-based and vision transformer-based) for detecting parking lots using semantic segmentation on satellite images. 
    \item We demonstrate that using images with Near-Infrared (NIR) channels (in addition to RGB) improves segmentation accuracy.
    \item We propose a post-processing pipeline that improves predictions by removing holes, simplifying edges, and utilizing publicly-available datasets to correct misclassified buildings and roads. These techniques further improve accuracy.
\end{itemize}

\section{Related Work}

Segmenting parking lots should be distinguished from other parking-related computer vision tasks that have received more attention in the literature. For example, \cite{Amato2016, Amato2017, Hung2022, Yuldashev2023} focus on measuring \emph{parking occupancy}  (i.e., whether spaces hold parked cars) by drawing bounding boxes around parking spaces or parking blocks \cite{Hurst-Tarrab2020, Vadivel2020} and then identifying parked cars. These tasks fall under \emph{object detection}, whereas our approach employs \emph{semantic segmentation} to detect the entire parking lot (see figure \ref{fig:annotation_ex} for distinctions between parking lots, spaces, and ``blocks''). Yin et al. \cite{Yin2022} is the only work that attempts to segment parking lots. While that study also utilizes contextual features such as roads, and buildings as input in the deep-learning model, we instead utilize such features in post-processing. Moreover, their dataset consists of 1,344 images\footnote{The images in Grab-Pklot \cite{Yin2022} are of higher resolution 1024 x 1024, which is equivalent to 5,376 images of 512 x 512} in Singapore, whereas ours consists of 12,617 images from the United States.

As \cite{Yin2022} point out, segmenting parking lots poses unique challenges: (i) variation in size and shape of parking lots; (ii) overlap with other objects such as vehicles and vegetation. In addition, since our goal is to segment parking lots across the United States, we also face (iii) substantial differences among cities in foliage, paving materials, sidewalk designs, and other factors.

Table \ref{tab:dataset} provides a summary of other datasets that are related to detecting parking lots.

\begin{table*}[h!]
    \centering
    \begin{tabular}{|l|c|c|c|c|c|}
        \hline
        \textbf{Dataset}                    & \textbf{Annotations} & \textbf{\# Images} & \textbf{View} & \textbf{Open-Source} & \textbf{NIR channel} \\ \hline
        % PKLot \cite{deAlmeida2015}   & Parking Space  & BB & 12,417 & Camera & \ding{51} & \ding{55} \\ \hline
        % CNRPark \cite{Amato2016}  & Parking Space  & BB & 12,000 & Camera & \ding{51} & \ding{55} \\ \hline
        % CNRPark+EXT \cite{Amato2017} & Parking Space  & BB & 144,965 & Camera & \ding{51} & \ding{55} \\ \hline
        % APKLOT \cite{Hurst-Tarrab2020}   & Parking Block  & BB & 500 & Satellite & \ding{51} & \ding{55} \\ \hline
        % Vadivel et al. \cite{Vadivel2020} & Parking Block  & BB & 1,200 & Satellite & \ding{55} & \ding{55} \\ \hline
        Safe Graph \cite{SafeGraph2024}     & Parking Lot          & NA                 & Satellite     & \ding{55}            & \ding{55}            \\ \hline
        Earth Define \cite{EarthDefine2021} & Parking Lot          & NA                 & Satellite     & \ding{55}            & \ding{55}            \\ \hline
        Grab-Pklot \cite{Yin2022}           & Parking Lot          & 1,344              & Satellite     & \ding{51}            & \ding{55}            \\ \hline
        ParkSeg, Ours                                & Parking Lot          & 12,617             & Satellite     & \ding{51}            & \ding{51}            \\ \hline
    \end{tabular}
    \caption{Comparison of datasets.}\
    \label{tab:dataset}
\end{table*}

\section{Dataset Construction}

This section describes the construction of the dataset, which consists of:

\begin{itemize}
    \item 297.7 \si{km^2} of total area
    \item 62.5 \si{km^2} of labeled parking area
    \item 35,127 annotated parking boundaries.
    \item 12,617 PNG image-mask pairs
\end{itemize}

\subsection{Parking Lot Annotations}

The first step in constructing the training dataset is to produce parking lot annotations. To produce these, we started out with two data sources:
\begin{itemize}
    \item Parking Reform Network (PRN) data: parking lots located in the downtown areas of 42 US cities. For most cities, this consists of the areas within the innermost beltway of freeways. These were created by the Parking Reform Network (PRN) for their interactive maps.
    \item OpenStreetMap (OSM) data downloaded for the entirety of three cities: Champaign IL, Anaheim CA, and Lubbock, TX. These cities were selected because they are in different parts of the country.
\end{itemize}
% \st{These data were originally GeoJSON files with parking lots as geometry polygons.}

\begin{figure}[ht]
    \centering
    \begin{subfigure}{0.23\textwidth}
        \centering
        \includegraphics[width=0.9\textwidth]{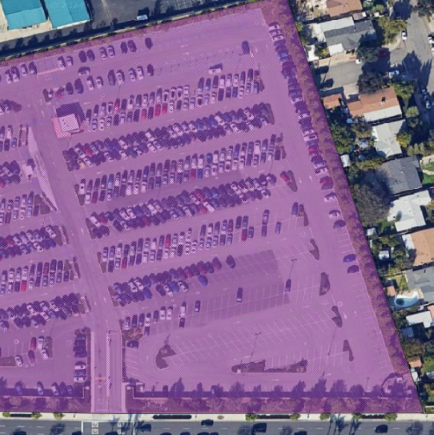}
        \caption{OSM Annotation}
        \label{fig:osm_original}
    \end{subfigure}
    \hfill
    \begin{subfigure}{0.23\textwidth}
        \centering
        \includegraphics[width=0.9\textwidth]{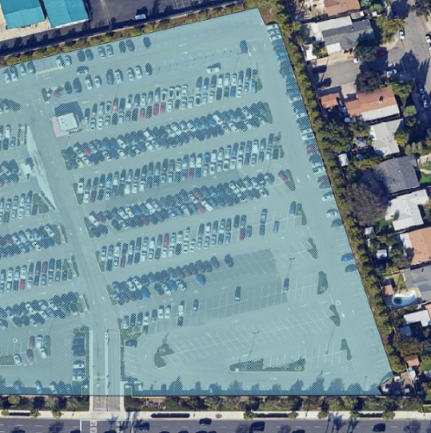}
        \caption{Corrected Annotation}
        \label{fig:osm_corrected}
    \end{subfigure}
    \caption{Corrected example with annotations drawn along pavement edge}
    \label{fig:borders}
\end{figure}

\begin{figure}[ht]
    \centering
    \begin{subfigure}{0.23\textwidth}
        \centering
        \includegraphics[width=0.9\textwidth]{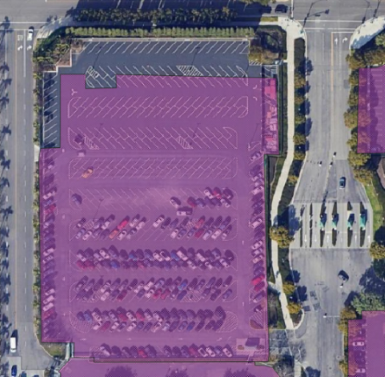}
        \caption{OSM Annotation}
        \label{fig:osm_original}
    \end{subfigure}
    \hfill
    \begin{subfigure}{0.23\textwidth}
        \centering
        \includegraphics[width=0.9\textwidth]{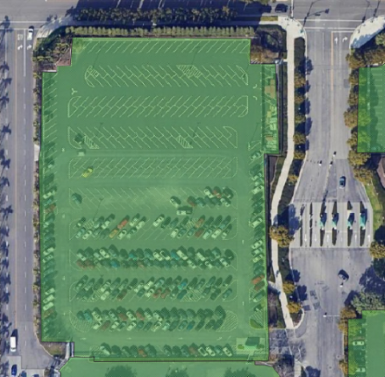}
        \caption{Corrected Annotation}
        \label{fig:osm_corrected}
    \end{subfigure}
    \caption{Corrected example where OSM annotation is out-of-sync with satellite image}
    \label{fig:osm_ex}
\end{figure}

Both datasets required substantial modification for the purpose of training. As \cite{Yin2022} notes, the OpenStreetMap (OSM) dataset contains many mistakes. More fundamentally, neither dataset was \emph{created} to train a model for semantic segmentation, which mattered in three primary ways:
\begin{enumerate}
    \item \emph{Garages:} Our dataset only includes parking garages with parking lots \emph{visibly} on top of them. The model cannot visually distinguish the purpose of a building.
    \item \emph{Boundaries:} Parking lot annotations must be drawn \emph{along the edge of the pavement,} in order for the deep learning model to learn visual cues. See Figure \ref{fig:borders} for an example. Within OSM, the annotations run along the edge of the \emph{parcel} rather than the parking lot itself.
    \item \emph{Outdated:} Annotations within OSM are not up-to-date at times and therefore are out-of-sync with the latest satellite images.  See Figure \ref{fig:osm_ex} for example, where the parking lot edges are drawn with reference to a building that no longer exists.
\end{enumerate}

In addition, for the 42 cities with PRN data, we expanded the dataset by adding several parking lots outside the downtown areas that PRN originally labeled.

To correct and expand the datasets, a team of students refined the data in the QGIS app by overlaying the polygons over a Google Satellite basemap. The correction process involved removing incorrect polygons that are no longer parking areas, adding missing parking polygons, and aligning the parking lots' edges to the Google basemap we are using. In the end, this process yielded the corrected annotations, which are exported from the QGIS app and partitioned into non-overlapping PNG images of 512x512 pixels. These label images (masks) are single-channel images with pixel values of `0' for the background and `1' for the parking areas.

Note that in drawing annotations, there are inevitably \emph{judgement calls} about what ``counts'' as parking. If a driveway leads to a parking lot, is that part of the parking? Or is it a private road which provides vehicle access from the street to a door that opens into the parking lot? For driveways, we have allowed students to rely on case-by-case judgement and encouraged them to incorporate only very short driveways---in order to keep the model from learning to recognize roads generally. Similarly, consider our decision to end parking at the edge of the pavement surface rather than the edge of a parcel. In understanding certain impacts of parking, it may be sensible to consider entire parcels when they are primarily used for parking. For example, if the parcel in Figure \ref{fig:borders} were not devoted to parking, then perhaps it could be a residential parcel yielding a certain expected property tax per acre. Different definitions may be better for some purposes than others.

\subsection{RGB Satellite Tiles}

\begin{figure}[ht]
    \centering
    \begin{subfigure}{0.15\textwidth}
        \centering
        \includegraphics[width=0.9\textwidth]{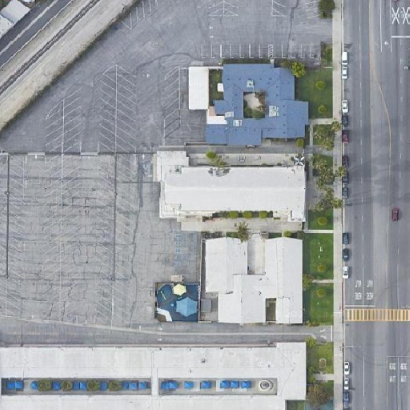}
        \caption{Image}
        \label{fig:image}
    \end{subfigure}
    \begin{subfigure}{0.15\textwidth}
        \centering
        \includegraphics[width=0.9\textwidth]{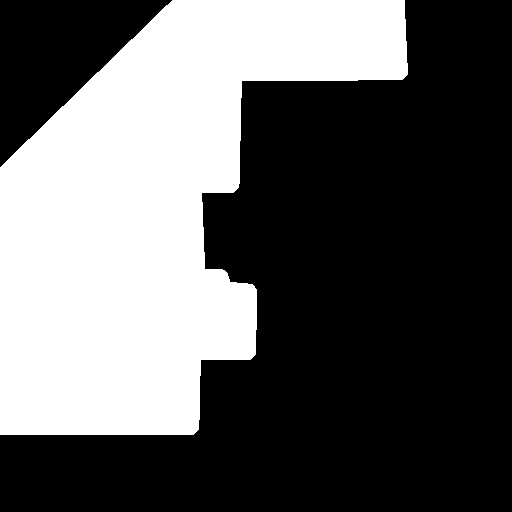}
        \caption{Mask}
        \label{fig:mask}
    \end{subfigure}
    \begin{subfigure}{0.15\textwidth}
        \centering
        \includegraphics[width=0.9\textwidth]{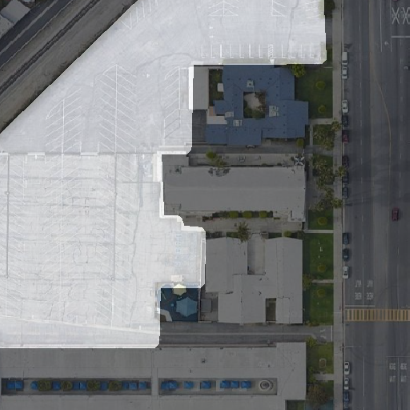}
        \caption{Mask and Image}
        \label{fig:mask_on_image}
    \end{subfigure}
    \caption{An example of a 512x512 pixels image and its annotation mask in our dataset}
    \label{fig:data_example}
\end{figure}

After the annotation masks were created, for basemap, we exported the Google Map tiles from QGIS to PNG files with a resolution of 30 cm per pixel, a size of 512x512 pixels each and three Red/Green/Blue color channels. Figure \ref{fig:data_example} shows an example image and its parking lot mask. There are no images in the dataset without parking areas. On average, 21\% of each image is covered by parking pixels.

\subsection{Enriching tiles with near-infrared}
% The complexity of the task, especially in detecting the edges of parking lots, requires us to enhance the information that the dataset can offer to the model. Therefore, we use an approach to enrich the data in hopes of improving the model's accuracy. This method includes adding an infrared channel to the 3-channel images as an additional channel.

% \subsubsection{Infrared Channel}

\begin{figure}[ht]
    \centering
    \begin{subfigure}{0.15\textwidth}
        \centering
        \captionsetup{labelformat=empty}
        \includegraphics[width=0.9\textwidth]{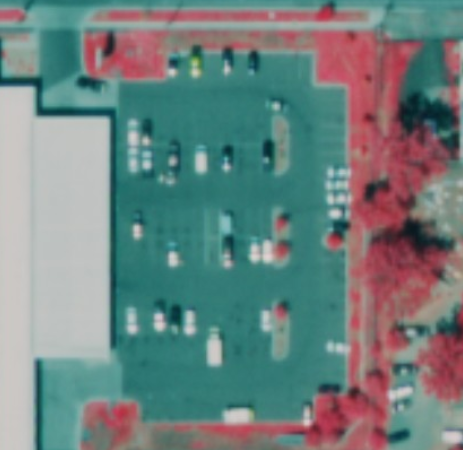}
        \caption{}
        \label{fig:naip_ex_a}
    \end{subfigure}
    \begin{subfigure}{0.15\textwidth}
        \centering
        \captionsetup{labelformat=empty}
        \includegraphics[width=0.9\textwidth]{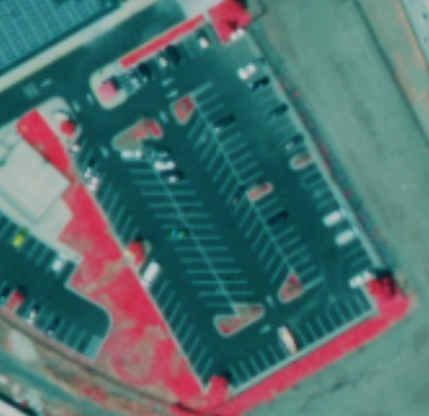}
        \caption{}
        \label{fig:naip_ex_b}
    \end{subfigure}
    \begin{subfigure}{0.15\textwidth}
        \centering
        \captionsetup{labelformat=empty}
        \includegraphics[width=0.9\textwidth]{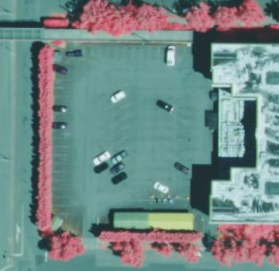}
        \caption{}
        \label{fig:naip_ex_c}
    \end{subfigure}
    \caption{Examples of NAIP imagery}
    \label{fig:naip_ex}
\end{figure}

The National Agriculture Imagery Program (NAIP) provides a tile set of aerial imagery for the United States which has \emph{four} color channels: Red, Green, Blue, and Near Infrared (NIR). Figure \ref{fig:naip_ex} shows some examples of this imagery, displayed with the true Red channel swapped for the NIR channel. Vegetation reflects more NIR, so the grass and foliage around the border of the parking lots stand out. We hypothesized that this contrast could aid the model in detecting edges, since many parking lots are surrounded by grass.

% After the satellite tiles are saved, we enrich them with a fourth infrared color channel. Infrared satellite or aerial imagery uses infrared wavelengths to detect heat emitted by objects. Most obviously, since parking surfaces are often surrounded by grass, foliage or curbs (which absorb heat less easily than asphalt), infrared improves contrast along edges. Perhaps less obviously, infrared signatures are less affected by temporary shadows than those of visible light, because it takes time for temperatures to change.
% For example, suppose that a satellite image is taken at 2 PM, when a tree casts a dark shadow over the edge of a parking lot. If the edge were in the sun all day prior, then the temperature difference between the parking and its environs will remain significant for hours.
% \begin{itemize}
%     \item The infrared channel improves the contrast between parking surface and its environs, thereby making it easier for the model to detect edges.
%     % \item Obstacles such as trees have significantly different infrared values, helping the model better understand the labels.
%     % \item In cases where there is a temperature difference between streets, buildings, and parking lots, an infrared channel is beneficial.
% \end{itemize}

% \begin{figure}[ht]
%     \centering
%     \includegraphics[width=0.4\textwidth]{figures/Naip_example.png}
%     \caption{An example of NAIP imagery}
%     \label{fig:naip_ex}
% \end{figure}

% \begin{figure}[ht]
%     \centering
%     \begin{subfigure}{0.3\textwidth}
%         \centering
%         \includegraphics[width=0.9\textwidth]{figures/Naip_example.png}
%         \caption{}
%         \label{fig:naip_ex_a}
%     \end{subfigure}
%     \begin{subfigure}{0.3\textwidth}
%         \centering
%         \includegraphics[width=0.9\textwidth]{figures/Naip_example_5.png}
%         \caption{}
%         \label{fig:naip_ex_b}
%     \end{subfigure}
%     \begin{subfigure}{0.3\textwidth}
%         \centering
%         \includegraphics[width=0.9\textwidth]{figures/Naip_example_3.png}
%         \caption{}
%         \label{fig:naip_ex_c}
%     \end{subfigure}
%     \caption{An example of NAIP imagery Extra (will be deleted, just to choose between images)}
%     \label{fig:naip_ex_extra}
% \end{figure}

% Ostensibly, one could use the NAIP tiles rather than Google's Satellite imagery
% One question we faced in 

A drawback to the NAIP dataset is that it has a resolution as low as 1 m/pixel in some areas, whereas the Google satellite imagery has 30 cm/pixel. Therefore, we created a \emph{second} training dataset of 30 cm/pixel tiles which combine the Red/Green/Blue channels from Google Maps with a ``resampled'' NIR channel exported from NAIP tiles. To convert the NIR channel to 30 cm/pixel, it is necessary to apply \emph{raster resampling}, which fill in missing pixel values. For resampling, we use \emph{bilinear interpolation}, which takes a weighted average of the four nearest pixels in the original image to determine the value of each missing pixel (see \cite{Kirkland2010}).

% Figure \ref{fig:bilinear_interpolation} illustrates this method. The value for the pixel at coordinates (x, y) is calculated using the pixel values of the four nearest points in the original image, with the weights equal to areas shown in the figure. XXX why are the areas different sizes? WHY THE COLORS DIFFERENT?

% \begin{figure}[ht]
%     \centering
%     \includegraphics[width=0.6\textwidth]{figures/bilinear_interpolation.png}
%     \caption{Bilinear interpolation method for resampling raster data}
%     \label{fig:bilinear_interpolation}
% \end{figure}

% A potential pitfall of this approach was that the NAIP data was sometimes slightly different angles than the Google Maps satellite data. XXX EXplainXXX EXplain

As a result, the dataset contains of \emph{two} sets of 12,617 image-mask pairs: one set includes 3-channel images, and the other set 4-channel images. Both sets have a 30 cm/pixel resolution and are 512x512 pixels.

\section{Method}
\begin{figure}[ht]
    \centering
    \includegraphics[width=0.5\textwidth]{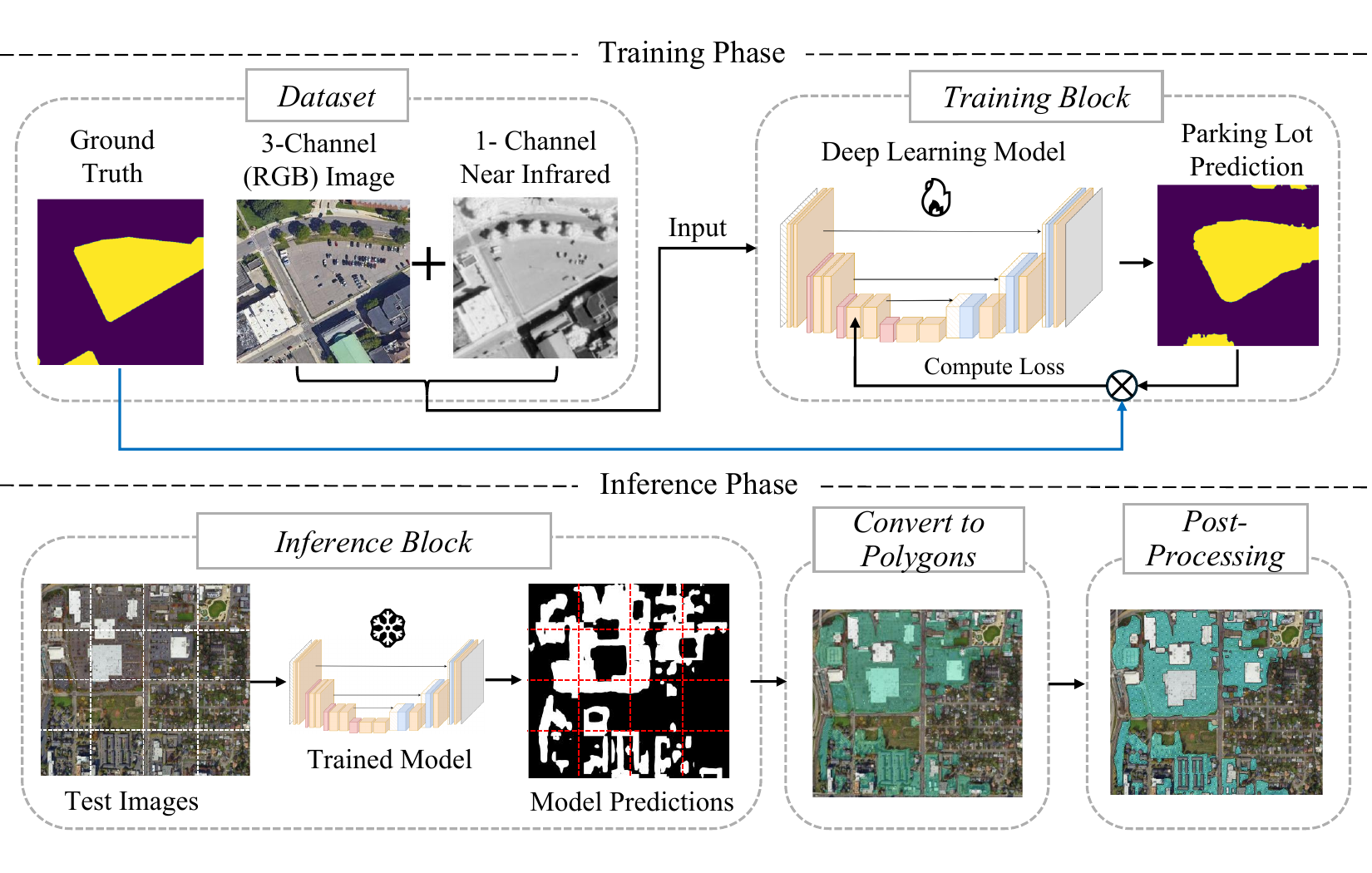}
    \caption{Workflow Diagram}
    \label{fig:workflow}
\end{figure}

This section describes the experimental procedure, including the deep learning models and the post-processing steps implemented to improve the accuracy of the results. Figure \ref{fig:workflow} shows the overall workflow, which is divided into two phases: training and inference. The training phase comprises two parts of data construction and model training. The inference phase includes obtaining predictions from the deep learning model and performing post-processing on the results.

\subsection{Deep Learning Models}

% Deep learning models are known to be particularly effective at processing and analyzing images. Their deep network architectures enable them to automatically extract complex features and patterns from images, learning to perform tasks such as classification, object detection, and segmentation. This is achieved by adjusting weights within the network that capture essential information for the given task. The process of determining these weights involves the use of a gradient descent algorithm, which is guided by a loss function that quantifies the penalty for errors made by the model.

We tackle the parking lot segmentation problem using \emph{semantic segmentation}: a core task in computer vision where each pixel in an image is classified into a predefined category. In our case, the task involves binary classification: identifying pixels as either ``parking'' or ``not parking.'' Today, Convolutional Neural Networks (CNNs) and vision transformers are two prevalent methods used for this task. Below, we test five models: two CNN-based, including Fully Convolutional Networks (FCNs) \cite{Long2015} and DeepLabV3 \cite{Chen2017}, and three transformer-based, including SegFormer \cite{Xie2021}, Mask2Former \cite{Cheng2022}, and OneFormer \cite{Jain2022}.

We utilize the models with their original architecture. Therefore, we do not delve into the architectures and layers here. The only modifications we make are to the first layer (when using 4-channel images as input) and the last layer, since we have only one class in addition to the background. The rest remains the same as in their original papers. The following sections will explain some key features of each model, and the pretrained weights we used for each one.
% Our problem is to identify the entire area of a parking lot .Therefore, a semantic segmentation task can provide the necessary information. In segmentation, we cluster parts of an image that belong to a specific class. In this case, our only class is the parking lot. Since segmentation involves predicting a category for each pixel, it essentially performs classification for every pixel rather than for the whole image. \citet{Long2015} was the first study to systematically demonstrate this capability for Convolutional Networks.

\subsubsection{FCN}
% Fully Convolutional Neural Networks (FCNs) \cite{Long2015} were the first models designed to classify each pixel of an image into a specific class using CNNs, allowing them to capture spatial hierarchies throughout the layers. We use an FCN model with pretrained weights of ResNet50. These pretrained weights provide a strong starting point for the model by leveraging features learned from a large and diverse dataset, leading to quicker convergence and enhanced performance when fine-tuning on our dataset.

Fully Convolutional Neural Networks (FCNs) \cite{Long2015} was the first architecture to convert classification-based CNNs into fully convolutional models that predict dense output for every pixel. This feature helps them to maintain spatial hierarchies throughout the network, allowing them to handle arbitrary input image sizes.

% Fully Convolutional Neural Networks (FCNs) \cite{Long2015} are poineering models for semantic segmentation task, designed to classify each pixel of an image into a specific class. Unlike traditional CNNs, which consider a single class for the entire image, this models generate a label map with the same dimentions as the image. The main innovation of this model is that it replaces the fully connected layers to convolutional layers, helping the model to maintain spatial hierarchies throughout the layers.

% We utilize the model with its original architecture. Therefore, we do not delve into the architecture and layers here. The only modifications we make are to the first layer (when using 4-channel images as input) and the last layer, since we have only one class. The rest remains the same as in the original paper. Moreover, we use FCN model with pretrained weights of ResNet101. These pretrained weights provide a solid starting point by leveraging features learned from a large and diverse dataset, leading to quicker convergence and enhanced performance when fine-tuning on our dataset.

\subsubsection{DeepLabV3}

% We use DeepLabV3 \cite{Chen2017}, an improved version of the well-known DeepLab model \cite{Chen2018a}. This model also uses CNNs along with some advanced techniques to gather features at different resolutions. We use the ResNet50 pre-trained weights for this model as well.

DeepLabV3 \cite{Chen2017}, an extension of the original DeepLab model \cite{Chen2018}, designed to address the challenge of capturing multi-scale contextual information in CNN-based approaches. By incorporating atrous (dilated) convolutions and the Atrous Spatial Pyramid Pooling (ASPP) module, DeepLabV3 effectively expands its receptive field and aggregates multi-scale features without increasing computational costs. 

% DeepLab \cite{Chen2018a} is a state-of-the-art deep learning model designed for semantic segmentation tasks using convolutional neural networks. In this paper, we use an improved version called DeepLabV3, introduced by the same group of researchers \cite{Chen2017}. This model employs advanced techniques such as atrous convolution, which helps capture multi-scale contextual information without losing spatial resolution, making it particularly effective for our task.

% Atrous (or dilated) convolutions \cite{Chen2018a} is one of the key features of this model allowing it to control the resolution at which feature responses are computed. This helps the model to have a larger field of view and capture more contexual information without increasing the number of parameters. Moreover, this model uses Atrous Spatial Pyramid Pooling (ASPP) module \cite{Chen2018a} which leverages atrous convolutions with different rates to examine the input image at multiple scales. This technique gathers features at different resolutions, helping te model better understand the objects at different scales.

% As with the FCN, the model architecture will be the same as the original paper, except for the first and last layers. Additionally, we use pretrained weights to help the model learn faster. In our case, we use ResNet50 weights available in PyTorch, which have shown better performance.

\subsubsection{SegFormer}
SegFormer \cite{Xie2021} is a novel deep learning model that combines the strengths of CNNs and transformers to segment images. Transformers, originally designed for natural language processing tasks, are capable of capturing long-range dependencies and contextual information. By combining this ability with the robust spatial feature extraction of CNNs, SegFormer excels in understanding both local details and global context.

\subsubsection{Mask2Former}
Mask2Former \cite{Cheng2022} adopts the same universal architecture as MaskFormer \cite{Cheng2021} but uses masked-attention instead of standard cross-attention used in transformer-based architectures. Masked attention only attends to the foreground region of the predicted mask for each query, which leads to a faster convergence and performance. They also implement multi-scale high-resolution features to handle small objects or regions. Unlike previous universal architectures, Mask2Former outperforms specialized architectures trained on specific tasks such as semantic or instance segmentation.

\subsubsection{OneFormer}
OneFormer \cite{Jain2022} is built on the idea to have a universal architecture and model for semantic, instance and panoptic segmentation tasks. Unlike Mask2Former, Oneformer does not require training on each task individually --- by employing a task-conditioned joint training strategy. They compute a query-text contrastive loss which helps the model learn inter-task distinctions. However, since our interest is semantic segmentation alone, we do not use contrastive loss while fine-tuning the model.

\subsubsection{Training Setup}

The training and validation sets are randomly partitioned, with a 90\%-10\% ratio. The test set, however, consists of 400 images from a different city that was not included in the training or validation phases (in addition to the 12,617 images mentioned).

All five models were implemented in Python using the PyTorch library \cite{Paszke2019}, and the experiments were conducted on an NVIDIA RTX A4000 GPU with 16 GB of memory and an Intel Xeon CPU. 

% \st{To increase the amount of training data and improve model performance, we utilized several data augmentation methods offered by the PyTorch library, including Random Horizontal Flip, Random Rotation, Resize, and Color Jitter (for more explanation of methods, see \cite{Shorten2019}). These augmentation methods create slightly different images randomly, which helps prevent the model from overfitting, leading to better performance \cite{Perez2017}.}
% \begin{itemize}
%     \item Random Horizontal Flip: Horizontally flips the image with a 50\% probability.
%     \item Random Rotation: Rotates the image by an angle of 15 degrees.
%     \item Resize: Changes the size of the image by a factor of 1.15.
%     \item Random Crop: Crops the image at a random location.
%     \item ColorJitter: Randomly change the brightness, contrast, saturation and hue of the image.
%     \item Normalize: Normalizes the pixel values of the image by their mean and standard deviation for all the channels. For the first three channels, the ImageNet mean and standard deviation are used. For the infrared channel, the mean and standard deviation of this channel from our own dataset are used.
% \end{itemize}

% Hyperparameters: 
Regarding hyperparameters, the learning rate is set to 1e-5, and the Adam optimizer is utilized. We also use an early stopping method to halt the model once it converged. This method had a patience value of 10 epochs.

% , meaning if there is no improvement in the model's performance after 10 epochs, training is stopped.

To optimize model performance, we need a loss function that quantifies model's errors during the training process. Since our problem involves binary classification for each pixel (``parking'' or ``not parking''), we use Binary Cross Entropy (BCE) with Logits Loss, which combines a Sigmoid layer and BCE loss \cite{Bishop2024}. Equation \ref{eq:loss} shows the loss function:

\begin{equation}\label{eq:loss}
    \small
    L = \frac{1}{N} \sum_{n} \left[ -w_n \left( y_n \cdot \log \sigma(x_n) + (1 - y_n) \cdot \log (1 - \sigma(x_n)) \right) \right],
\end{equation}
where $L$ is the loss value, $N$ is the batch size, $w_n$ is a weight for positive class, $y_n$ is the true label of sample $n$, $x_n$ is the input feature of sample $n$, and $\sigma$ denotes the sigmoid function.

% \begin{equation}\label{eq:sigmoid}
%     \sigma(x) = \frac{1}{1+e^{-x}}
% \end{equation}

Since the proportions of background and parking lots in the images are significantly different (on average, 21\% parking and 79\% background), we have an imbalanced class situation. Therefore, we use $w_n = \frac{1}{0.21}$ as the positive weight in the loss function. This assigns a factor of 4.76 to the parking labels (positive class) and 1 to the background, prioritizing the correct prediction of parking lot pixels over background pixels.

% Evaluation method: 
Model performance is evaluated using two commonly-used metrics:
\begin{itemize}
    \item \textbf{Pixel-wise accuracy (PW):} Pixel-wise accuracy measures the percentage of pixels that are correctly predicted as either background or parking lot.
    \item \textbf{mean Intersection over Union (mIoU):} This metric evaluates the average IoU for all classes (including background), measuring the overlap between the predicted mask and the ground truth mask:

          \begin{equation}\label{eq:Jaccard}
              IoU(A,B) = \frac{|A \cap B|}{|A \cup B|}
          \end{equation}

          where $|A \cap B|$ is the area of overlap (intersection) between the predicted and ground truth masks, and $|A \cup B|$ is the area of union between the predicted and ground truth masks. This metric is computed for all the classes and the average is called mIoU.
\end{itemize}

\subsection{Post-Processing}

% The output
Our \emph{post-processing} stage modifies the outputs of the deep learning models, in order to obtain edges that are simpler and more accurate. To do so, we first convert the prediction masks into a single GeoJSON file containing the predicted parking polygons. Since we know the geographic bounding box of every image in the test dataset, it is straightforward to map every pixel in each test to a latitude/longitude pair. After doing so, we render all the predicted parking lot annotations as geometry polygons in a single, large GeoJSON file. Once the GeoJSON file is created, we carry out our post-processing tasks by modifying the polygons. These tasks include (i) removing small ``holes'', (ii) simplifying boundary edges, (iii) removing areas which are actually buildings, and (iv) removing areas which are roads. The following sections explain each task in more detail.

\subsubsection{Removing Holes}

\begin{figure}[ht]
    \centering
    \begin{subfigure}{0.23\textwidth}
        \centering
        \includegraphics[width=0.9\textwidth]{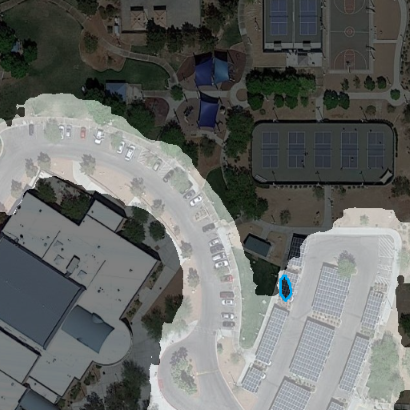}
        \caption{Example of gap in the mask}
        \label{fig:hole_in}
    \end{subfigure}
    \begin{subfigure}{0.23\textwidth}
        \centering
        \includegraphics[width=0.9\textwidth]{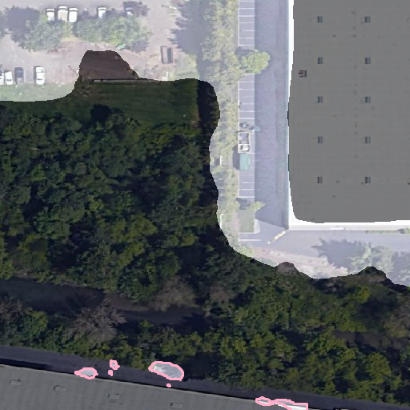}
        \caption{Example of mask holes}
        \label{fig:hole_out}
    \end{subfigure}
    \caption{An example polygon before and after removing holes}
    \label{fig:hole_ex}
\end{figure}

The deep learning models often output masks with small ``holes,'' which are nearly always incorrect. There are two kinds of holes. Figure \ref{fig:hole_in} shows a hole (outlined in blue) which is a small, erroneous gap in the parking lot mask. Figure \ref{fig:hole_out} shows holes (in pink) which are small masks themselves. Manual inspection revealed that nearly all such holes are errors in output. While parking lots vary in size and sometimes have ``islands'' within them, it is apparently the case that, the smaller the hole, the more likely it is to be an error. Therefore, we hypothesized that eliminating holes smaller than a defined threshold would improve the model's accuracy. By trial-and-error, we chose 60 $m^2$ as the threshold.

\subsubsection{Simplifying Edges}

Rough and complex edges are another common issue in segmentation output. Figure \ref{fig:hole_ex} also illustrates this problem. Rough edges often indicate inaccurate predictions, because parking lots have relatively simple edges. Rough edges also make it more difficult for someone to manually correct the model output later, since the resulting polygons have more vertices to fix.

\begin{figure}[ht]
    \centering
    \begin{subfigure}{0.23\textwidth}
        \centering
        \includegraphics[width=0.9\textwidth]{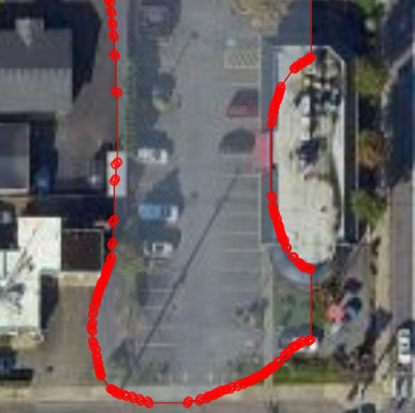}
        \caption{Before simplification}
        \label{fig:simp_before}
    \end{subfigure}
    \begin{subfigure}{0.23\textwidth}
        \centering
        \includegraphics[width=0.9\textwidth]{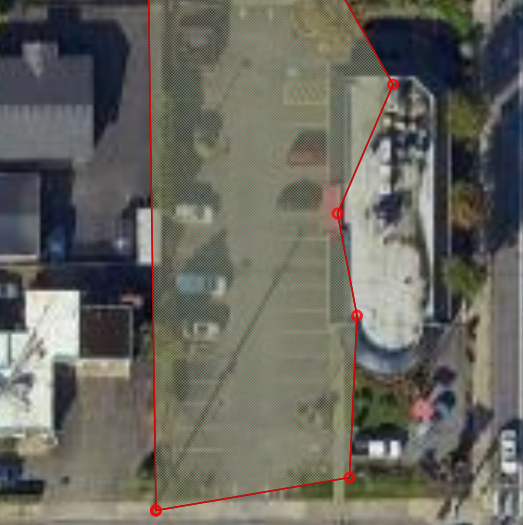}
        \caption{After simplification}
        \label{fig:simp_after}
    \end{subfigure}
    \caption{An example polygon before and after simplification}
    \label{fig:simplification_ex}
\end{figure}

Thus, our next post-processing step is to simplify the parking lot edges. To do so, we use Mapshaper \cite{Harrower2006}: a tool for editing Shapefile, GeoJSON, and raster data formats. We use the tool to apply the the Douglas-Peucker algorithm for shape simplification \cite{Douglas1973}. The goal of this algorithm is to reduce the total number of points composing a line or curve while maintaining a similar shape. We implement this algorithm in Python by setting a maximum percentage of points that can be removed. Figure \ref{fig:simplification_ex} shows an example of a parking lot annotation, along with the vertices before and after simplification. The simplified polygon classifies fewer non-parking pixels as parking. It is also much easier to correct manually in GIS software by dragging vertices into place.

\subsubsection{Removing Buildings}

\begin{figure}[ht]
    \centering
    \begin{subfigure}{0.23\textwidth}
        \centering
        \includegraphics[width=0.9\textwidth]{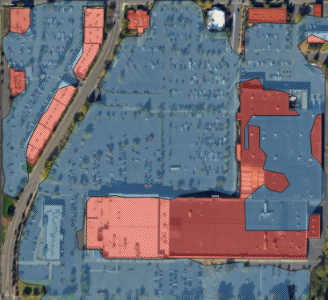}
        \caption{Before building correction}
        \label{fig:build_before}
    \end{subfigure}
    \hfill
    \begin{subfigure}{0.23\textwidth}
        \centering
        \includegraphics[width=0.9\textwidth]{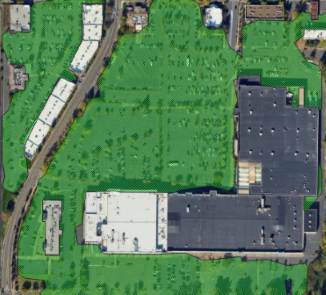}
        \caption{After building correction}
        \label{fig:build_after}
    \end{subfigure}
    \caption{An example polygon before and after correcting buildings' mistakes. The building dataset is displayed in red in (a).}
    \label{fig:building_ex}
\end{figure}

Building roofs are similar in color and shape to parking lots. Consequently, the models often misclassify parts of roofs as parking. To address this problem, Yin et al. \cite{Yin2022} incorporate extra channels containing building information into their model with success. We instead utilize the open-access ``building footprint'' dataset  provided by Microsoft \cite{Microsoft2018}, which provides annotations for all nearly all buildings in the United States. We take the model's prediction and simply subtract the building footprints using Python's Shapely library. Figure \ref{fig:building_ex} shows an example of an area before and after building removal.
% Building polygons are simply subtracting from the predicted parking polygons.

% The building polygons come from an open-access, computer-generated  \cite{Microsoft2018}. We remove the overlap between this dataset and the predicted parking data from our output, using the Shapely library in Python. 

\subsubsection{Removing Roads}
\begin{figure}[ht]
    \centering
    \begin{subfigure}{0.23\textwidth}
        \centering
        \includegraphics[width=0.9\textwidth]{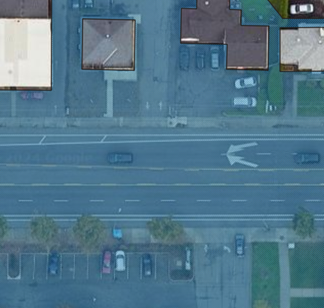}
        \caption{Before road correction}
        \label{fig:road_before}
    \end{subfigure}
    \hfill
    \begin{subfigure}{0.23\textwidth}
        \centering
        \includegraphics[width=0.9\textwidth]{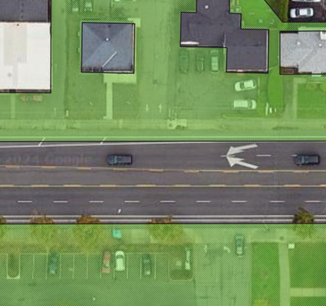}
        \caption{After road correction}
        \label{fig:road_after}
    \end{subfigure}
    \caption{An example polygon before and after correcting roads' mistakes.}
    \label{fig:road_ex}
\end{figure}

Roads are the most similar landscape feature to parking lots, so the models sometimes mistake portions of roads for parking. Yin et al. \cite{Yin2022} address this challenge by incorporating roadways as additional input channels in training data. For computational simplicity, we incorporate roadway information into post-processing. The road data is sourced from OpenStreetMap, where roads are represented as LineStrings that follow road centerlines. We create buffers around each centerline with widths that vary with the number of lanes recorded for each road. Then, the buffers are subtracted from the predicted parking lot polygons. Figure \ref{fig:road_ex} shows an example.

\begin{table*}[ht!]
    \centering
    \caption{Performance Comparison of Different Models. \textbf{Bold} indicates the best model for the metric and technique. \fbox{Box} indicates the best model overall for the respective metric}\label{tab:model_performance}
    \begin{tabular}{lcclccccccc}
        \toprule
        \multirow{2}{*}{Model} & \multirow{2}{*}{Backbone} & \multirow{2}{*}{\shortstack{Pre-training \\ Dataset}} & \multirow{2}{*}{\shortstack{Training \\ Images}} & \multicolumn{2}{c}{Original} & \multicolumn{2}{c}{\shortstack{w/ Building \\ Removal}} & \multicolumn{2}{c}{\shortstack{w/ Road \\ Removal}}  \\
        \cmidrule(lr){5-6} \cmidrule(lr){7-8} \cmidrule(lr){9-10}
        & & & & mIoU & PW & mIoU & PW & mIoU & PW \\
        \midrule
        \multirow{2}{*}{FCN} & \multirow{2}{*}{ResNet50} & \multirow{2}{*}{COCO \cite{Lin2014}} & RGB & 77.92 & 94.22 & 79.55 & 94.54 & 80.18 & 94.81 \\
        & & & RGB + NIR & 80.18 & 94.83 & 82.13 & 95.43 & 82.45 & 95.54 \\
        \midrule
        \multirow{2}{*}{DeepLabV3} & \multirow{2}{*}{ResNet50} & \multirow{2}{*}{COCO \cite{Lin2014}} & RGB & 79.62 & 94.76 & 81.56 & 95.31 & 81.89 & 95.43 \\
        & & & RGB + NIR & 80.06 & 94.91 & 82.32 & 95.48 & 82.74 & 95.63 \\
        \midrule
        \multirow{2}{*}{SegFormer} & \multirow{2}{*}{MiT-B0} & \multirow{2}{*}{ADE20K\cite{Zhou2017}} & RGB & 81.47 & 95.33 & 82.21 & 95.59 & 82.42 & 95.66 \\
        & & & RGB + NIR & 81.75 & 95.24 & 83.32 & 95.79 & 83.53 & 95.87 \\
        \midrule
        \multirow{2}{*}{Mask2Former} & \multirow{2}{*}{\shortstack{Swin-L}} & \multirow{2}{*}{CityScapes\cite{Cordts2016}} & RGB & 82.04 & 95.22 & 82.69 & 95.74 & 82.99 & 95.85 \\
        & & & RGB + NIR & 82.06 & 95.23 & 83.31 & 95.68 & 83.92 & 95.88 \\
        \midrule
        \multirow{2}{*}{OneFormer} & \multirow{2}{*}{\shortstack{Swin-L}} & \multirow{2}{*}{COCO \cite{Lin2014}} & RGB & 83.23 & 95.72 & 84.55 & 96.15 & 84.73 & 96.21 \\
        & & & RGB + NIR & $\mathbf{84.05}$ & $\mathbf{96.07}$ & $\mathbf{84.59}$ & $\mathbf{96.26}$ & \fbox{$\mathbf{84.86}$} & \fbox{$\mathbf{96.34}$} \\
        \bottomrule
    \end{tabular}
\end{table*}

\section{Results and Analysis}

This section presents the results of the five deep learning models, both before and after our post-processing steps. Table \ref{tab:model_performance} shows the accuracy results. Since the first two post-processing tasks—removing holes and simplifying edges—have minimal individual effects, they are evaluated jointly along with the building removal task. The results for this combined evaluation are reported under the category ``w/ Building Removal." Moreover, to simplify comparisons given the numerous metrics, the post-processing results are cumulative: for example the ``w/ Road Correction" results are calculated based on the ``w/ Building" output. The best performance was achieved using the OneFormer model trained by 4-channel images after post processing, with an mIoU of 84.4\% and pixel-wise accuracy of 96.2\%. More discussion of the results and examples appear below.

% \begin{table}[ht]
%   \centering
%   \caption{Performance Comparison of Different Models}\label{tab:model_performance_old}
%   \begin{tabular}{llcccccc}
%   \toprule
%   \multirow{2}{*}{Model} & \multirow{2}{*}{Training Images} & \multicolumn{2}{c}{Original} & \multicolumn{2}{c}{w/ Simplification} & \multicolumn{2}{c}{w/ Building Correction} \\
%   \cmidrule(lr){3-4} \cmidrule(lr){5-6} \cmidrule(lr){7-8}
%    &  & mIoU & pixel-wise & mIoU & pixel-wise & mIoU & pixel-wise \\
%   \midrule
%   \multirow{3}{*}{DeepLabV3} & RGB & 67.3 & 86.1 & 68 & 86.7 &  &  \\
%    & RGB + Infrared & 70.6 & 87.4 & 71.6 & 88 & 75.7 & 89.9 \\
%    & RGB + SAM & 62.9 & 82.6 & 64.3 & 83.9 & 67.3 & 86.1 \\
%   \midrule
%   \multirow{3}{*}{FCN} & RGB &  &  &  &  &  &  \\
%    & RGB + Infrared &  &  &  &  &  &  \\
%    & RGB + SAM &  &  &  &  &  &  \\
%   \midrule
%   \multirow{3}{*}{SegFormer} & RGB & 74.3 & 89.2 & 75.1 & 89.7 &  &  \\
%    & RGB + Infrared & 75.9 & 90.6 & 76.5 & 90.6 & 79.4 & 92.5 \\
%    & RGB + SAM &  &  &  &  &  &  \\
%   \bottomrule
%   \end{tabular}
% \end{table}

\subsection{Model Comparison}
Table \ref{tab:model_performance} shows OneFormer significantly outperforms the other models on both metrics. OneFormer achieves approximately an 5.3\% increase in the mIoU metric compared to FCN for the model trained with RGB images and an 3.9\% increase when the NIR channel is included. The table is organized by performance, from FCN to OneFormer, with higher-performing models listed later. SegFormer also shows an improvement, with a 3.6\% higher mIoU for RGB-trained models compared to FCN. This improvement is particularly notable given that SegFormer has only 3.7 million parameters compared to FCN's 49.6 million. The same situation
holds when comparing DeepLabV3 and SegFormer. Overall, models leveraging vision transformers demonstrated better performance than both DeepLabV3 and FCN.

\begin{figure}[ht]
    \centering
    \begin{subfigure}{0.195\textwidth}
        \centering
        \includegraphics[width=\textwidth]{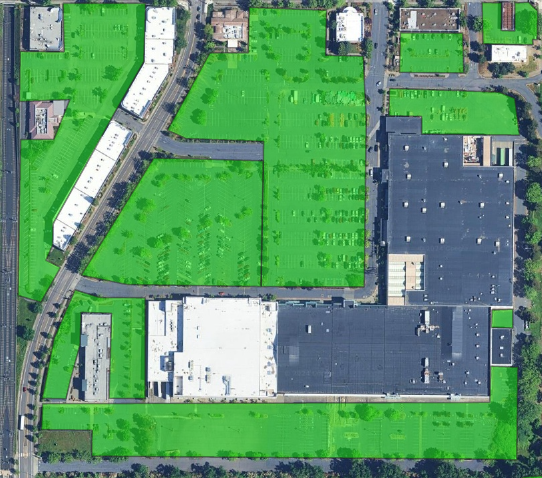}
        \caption{Ground Truth}
        \label{fig:ground_truth}
    \end{subfigure}
    \begin{subfigure}{0.195\textwidth}
        \centering
        \includegraphics[width=\textwidth]{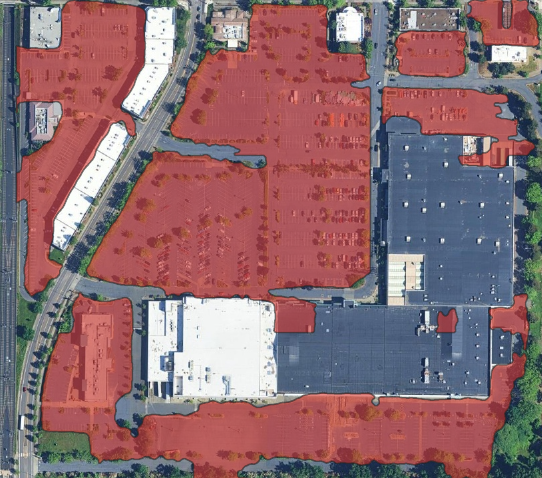}
        \caption{FCN}
        \label{fig:fcn_models}
    \end{subfigure}
    \begin{subfigure}{0.195\textwidth}
        \centering
        \includegraphics[width=\textwidth]{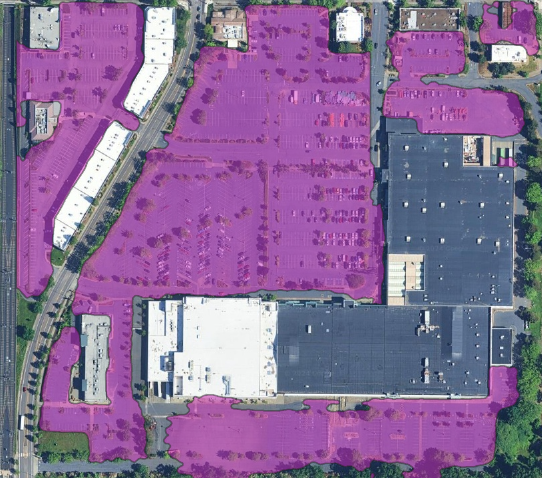}
        \caption{DeepLabV3}
        \label{fig:deeplab_models}
    \end{subfigure}
    \begin{subfigure}{0.195\textwidth}
        \centering
        \includegraphics[width=\textwidth]{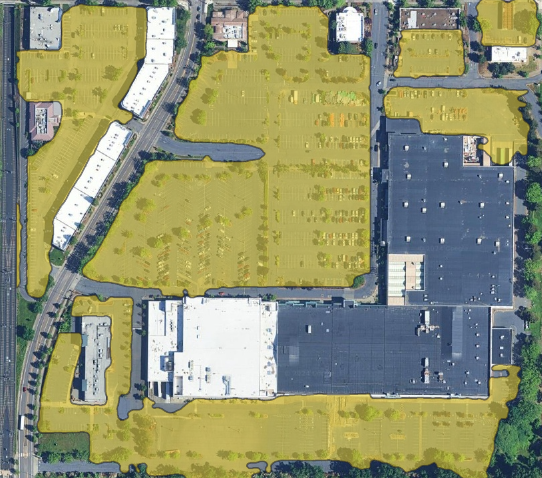}
        \caption{Segformer}
        \label{fig:segformer_models}
    \end{subfigure}
    \begin{subfigure}{0.195\textwidth}
        \centering
        \includegraphics[width=\textwidth]{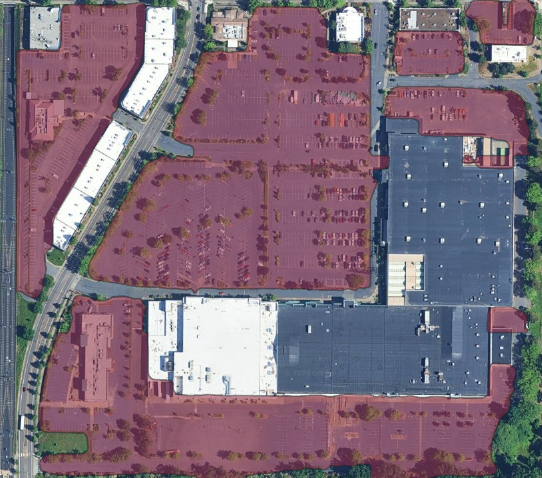}
        \caption{Mask2Former}
        \label{fig:mask2former_models}
    \end{subfigure}
    \begin{subfigure}{0.195\textwidth}
        \centering
        \includegraphics[width=\textwidth]{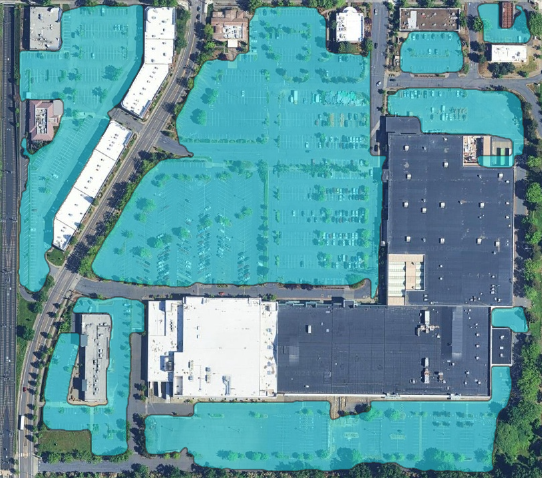}
        \caption{Oneformer}
        \label{fig:oneformer_models}
    \end{subfigure}
    \caption{Comparison between different models across a test area. The colored polygons indicate the true parking areas in (a), and the predictions in the remaining subfigures. The predictions are from models trained with RGB+NIR images and shown before post-processing.}
    \label{fig:models_comparison}
\end{figure}

% \begin{figure*}[ht]
%     \centering
%     \begin{subfigure}{0.24\textwidth}
%         \centering
%         \includegraphics[width=\textwidth]{figures/ground_truth_total_3.png}
%         \caption{Ground Truth}
%         \label{fig:ground_truth}
%     \end{subfigure}
%     \begin{subfigure}{0.24\textwidth}
%         \centering
%         \includegraphics[width=\textwidth]{figures/deeplab_RGB_build_total_3.png}
%         \caption{DeepLabV3}
%         \label{fig:deeplab_models}
%     \end{subfigure}
%     \begin{subfigure}{0.24\textwidth}
%         \centering
%         \includegraphics[width=\textwidth]{figures/fcn_RGB_build_total_3.png}
%         \caption{FCN}
%         \label{fig:fcn_models}
%     \end{subfigure}
%     \begin{subfigure}{0.24\textwidth}
%         \centering
%         \includegraphics[width=\textwidth]{figures/segformer_RGB_build_total_3.png}
%         \caption{Segformer}
%         \label{fig:segformer_models}
%     \end{subfigure}
%     \caption{Comparison between different models across the entire test area. The colored polygons indicate the true parking areas in (a), and the predictions in b and c. The predictions are from models trained with RGB images.}
%     \label{fig:models_comparison}
% \end{figure*}

Figure \ref{fig:models_comparison} provides a visual comparison of model performance, with subfigure (a) showing the "ground truth" and subfigures (b) through (f) displaying predictions from five different models for an example area of the test dataset. In this case, all models shown are only trained on RGB channels. (NIR enrichment is discussed below.) While all models demonstrate acceptable performance, DeepLabV3 and FCN tend to overpredict parking lot areas, mistakenly classifying non-parking pixels, such as parts of buildings or roads, as parking. In contrast, the predictions from the other models more closely align with the ground truth. Although post-processing tasks correct some of these errors, challenges in accurately detecting edges limits the overall accuracy for the first two models.

% \begin{figure}[ht]
%     \centering
%     \begin{subfigure}{0.15\textwidth}
%         \centering
%         \captionsetup{labelformat=empty}
%         \includegraphics[width=0.9\textwidth]{figures/ex_model_impact.png}
%         \caption{}
%         \label{fig:model_impact_ex_a}
%     \end{subfigure}
%     \begin{subfigure}{0.15\textwidth}
%         \centering
%         \captionsetup{labelformat=empty}
%         \includegraphics[width=0.9\textwidth]{figures/ex_model_impact_2.png}
%         \caption{}
%         \label{fig:model_impact_ex_b}
%     \end{subfigure}
%     \begin{subfigure}{0.15\textwidth}
%         \centering
%         \captionsetup{labelformat=empty}
%         \includegraphics[width=0.9\textwidth]{figures/ex_model_impact_3.png}
%         \caption{}
%         \label{fig:model_impact_ex_c}
%     \end{subfigure}
%     \caption{Examples of Segformer outperforming DeepLabV3: Yellow polygons show DeepLabV3 predictions, while purple polygons are from Segformer.}
%     \label{fig:model_impact_ex}
% \end{figure}

% Figure \ref{fig:model_impact_ex} shows a closer look at three cases where SegFormer outperforms DeepLabV3. As the figure shows, DeepLabV3 struggles with accurately detecting edges and classifies more non-parking areas as parking.

\subsection{Impact of Near-Infrared Channel}

Table \ref{tab:model_performance} shows that incorporating the NIR channel improved all models' accuracy. This improvement is most pronounced in FCN, with a 1.3\% increase in mIoU, compared to the other models. This suggests that the NIR channel has the most benefits when the model's baseline performance is lower.

\begin{figure}[ht]
    \centering
    \begin{subfigure}{0.149\textwidth}
        \centering
        \includegraphics[width=0.9\textwidth]{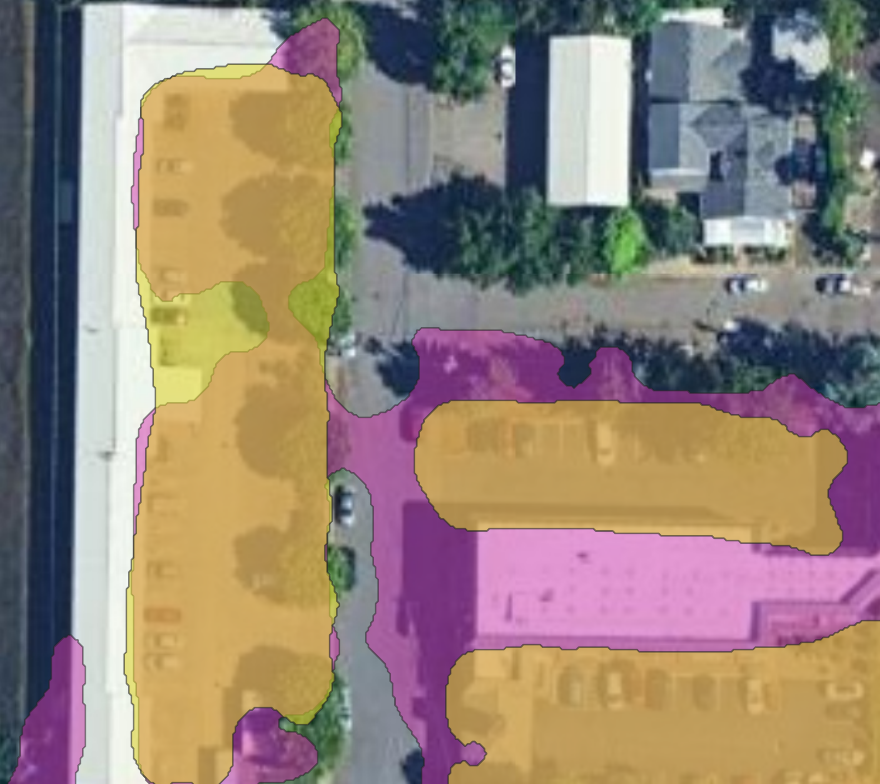}
        \caption{FCN}
        \label{fig:infra_impact_ex_a}
    \end{subfigure}
    \begin{subfigure}{0.153\textwidth}
        \centering
        \includegraphics[width=0.9\textwidth]{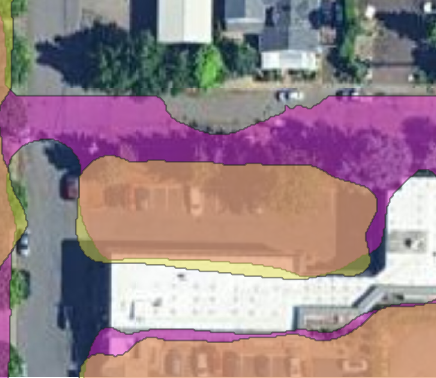}
        \caption{DeepLabV3}
        \label{fig:infra_impact_ex_b}
    \end{subfigure}
    \begin{subfigure}{0.15\textwidth}
        \centering
        \includegraphics[width=0.9\textwidth]{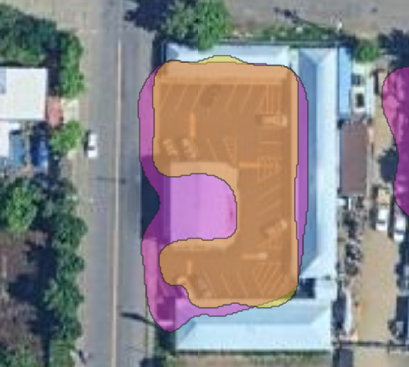}
        \caption{SegFormer}
        \label{fig:infra_impact_ex_c}
    \end{subfigure}
    \begin{subfigure}{0.15\textwidth}
        \centering
        \includegraphics[width=0.9\textwidth]{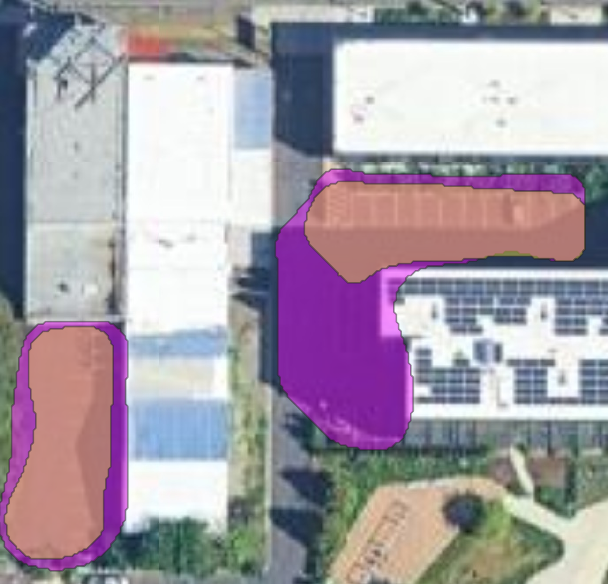}
        \caption{Mask2Former}
        \label{fig:infra_impact_ex_d}
    \end{subfigure}
    \begin{subfigure}{0.15\textwidth}
        \centering
        \includegraphics[width=0.9\textwidth]{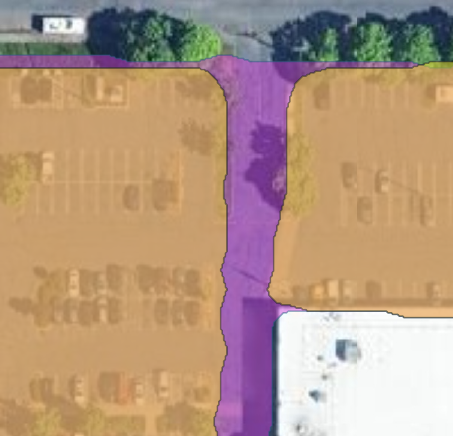}
        \caption{OneFormer}
        \label{fig:infra_impact_ex_e}
    \end{subfigure}
    \caption{Examples of impact of adding the NIR channel: Purple polygons show RGB model predictions; yellow shows RGB+NIR. The predictions are shown before post-processing.}
    \label{fig:infra_impact}
\end{figure}

Figure \ref{fig:infra_impact} shows five examples in which the NIR channel improves models performance. In this figure, we overlay the predictions of two models (one trained with RGB images and the other with RGB+NIR images) to better highlight the differences. It is clear that model with NIR detects edges better and overpredicts parking less often. In particular, the NIR channel keeps the model from erroneously merging nearby parking lots together, as seen in subfigures (a), (b), and (e). Reasonably, this happens because the NIR channel makes the grassy areas between nearby parking lots stand out more clearly. In subfigures (a) and (c) it aided in distinguishing buildings from parking lots, and in subfigure (a), it ensured the entire parking lot is recognized without missing any sections. Across all examples, the NIR channel consistently mitigates overprediction issues.

% The subfigures (a)-(d) are to be viewed in pairs without and with the NIR channel training: (a)/(b) show one area, (c)/(d) a second. For both pairs, it is clear that model with NIR detects edges better and overpredicts parking less often. In particular, the NIR channel keeps the model from erroneously merging nearby parking lots together. Reasonably, this happens because the NIR channel makes the grassy areas between nearby parking lots stand out more clearly.

% \begin{figure}[ht]
%     \centering
%     \begin{subfigure}{0.15\textwidth}
%         \centering
%         \includegraphics[width=0.9\textwidth]{figures/ex_infra_impact_deeplab.png}
%         \caption{}
%         \label{fig:infra_impact_ex_a}
%     \end{subfigure}
%     \begin{subfigure}{0.15\textwidth}
%         \centering
%         \includegraphics[width=0.9\textwidth]{figures/ex_infra_impact_deeplab_2.png}
%         \caption{}
%         \label{fig:infra_impact_ex_b}
%     \end{subfigure}
%     \begin{subfigure}{0.15\textwidth}
%         \centering
%         \includegraphics[width=0.9\textwidth]{figures/ex_infra_impact_segformer_4.png}
%         \caption{}
%         \label{fig:infra_impact_ex_c}
%     \end{subfigure}
%     \caption{Examples of impact of adding the NIR channel: PPurple polygons show RGB model predictions; green shows RGB+NIR. In (a) and (b), green overlays purple; in (c), purple overlays green.}
%     \label{fig:infra_impact_ex}
% \end{figure}

% Figure \ref{fig:infra_impact_ex} illustrates examples in a closer look where adding the NIR channel improved the model's performance. In this figure, we overlay the predictions of two models to better highlight the differences. In Figure \ref{fig:infra_impact_ex_a}, the NIR channel helped mitigate the issue of overprediction. In Figure \ref{fig:infra_impact_ex_b} it aided in distinguishing buildings from parking lots. Finally, in Figure \ref{fig:infra_impact_ex_c} it helped the model recognize the entire area as a parking lot, ensuring no parts were missed.

\subsection{Impact of Post-processing}
We analyze post-processing in two stages: building removal (which includes removing holes, smoothing edges, and correcting misclassified building polygons) and road removal. Table \ref{tab:model_performance} shows that building removal improved the models' mIoU accuracy by an average of 1.38\% and pixel-wise accuracy by 0.55\%, with a more pronounced effect in DeepLabV3 (2.1\% in mIoU) and FCN (1.8\% in mIoU). SegFormer, Mask2Former, and OneFormer saw smaller but still meaningful gains, given their already high baseline accuracies. The final step, road removal, further improved mIoU by 0.35\% and pixel-wise accuracy by 0.12\%. Despite their computational efficiency, these post-processing steps yield notable improvements, enhancing the overall quality of the predictions.

% We analyze post-processing in two stages: building removal (which includes removing holes, smoothing edges, and correcting misclassified building polygons) and road removal. Table \ref{tab:model_performance} shows that, on average, building removal improved the models' mIoU accuracy by 1.38\% and pixel-wise accuracy by 0.55\%, with a more pronounced effect observed in DeepLabV3 and FCN. Specifically, this stage resulted in an average mIoU increase of 2.1\% for DeepLabV3, 1.8\% for FCN, 1.1\% for SegFormer, and 0.95\% for both Mask2Former and OneFormer. While these improvements may seem modest, they are meaningful given the already high baseline accuracies. The final post-processing step, road removal, further enhanced performance, with an average mIoU increase of 0.35\% and a 0.12\% boost in pixel-wise accuracy. Despite their computational efficiency, these post-processing steps yield notable improvements, enhancing the overall quality of the predictions.

\section{Conclusion}
% City policies have mandated a minimum parking requirements for buildings which has become a controvertial topic recently. Opponents argue that these rules are unreasonable and have led to excessive parking with harmful consequences. Despite efforts by groups such as the Parking Reform Network and OpenStreetMap, there is a lack of comprehensive parking datasets to investigate the issue. 
This paper introduces a multi-step procedure for predicting parking lot segments from satellite images and producing GeoJSON files that users can easily correct. We constructed a dataset of two sets of 12,617 image-mask pairs: one with 3-channel images (RGB) and another with 4-channel images (RGB + Near-Infrared). These images were used to train five deep learning models (OneFormer, Mask2Former, SegFormer, DeepLabV3 and FCN) for image segmentation. Four post-processing steps carried out on the predicted masks were shown to improve model accuracy, as did enriching images with an NIR channel---even though the NIR channel as upsampled to a higher resolution. The best performance was achieved using the OneFormer model trained on 4-channel images with post-processing, with a mIoU of 84.9\% and a pixel-wise accuracy of 96.3\%.

The fact that an automated process can achieve reasonably high accuracy shows a pathway for deep learning to contribute to current discussion of urban form and public policy. This process---or a similar one---could be used to produce statistics, plots and maps of interest to those considering changes to parking minimums or other laws.
\clearpage

    %%%%%%%%% REFERENCES
    {\small
        \bibliographystyle{ieee_fullname}
        \bibliography{parking-detection}
    }

\end{document}